\documentclass[runningheads]{llncs}
\usepackage[T1]{fontenc}
\usepackage{amsmath}%
\usepackage{todonotes}
\usepackage{amsfonts}
\usepackage{graphicx}
\usepackage{multicol}
\usepackage{wrapfig}
\usepackage{multirow}
\usepackage{todonotes}
\usepackage{floatrow}
\usepackage{subcaption}

\usepackage{tikz}
\usepackage{booktabs}       % professional-quality tables
\usepackage{graphicx}
\begin{document}
\title{Towards Onboard Continuous Change Detection for Floods}
% \title{HiT: History-Injection Transformers for Onboard Continuous Flood Change Detection}
%
\titlerunning{HiT}
% If the paper title is too long for the running head, you can set
% an abbreviated paper title here
%
\author{
% Anonymous
Daniel Kyselica\inst{1}%\orcidID{0000-1111-2222-3333} 
\and
Jon\'a\v{s} Herec\inst{1,2}%pridaj si univerzity nech sa ti to pocita
\and
Oliver Kutis\inst{1}
\and
Rado Pito\v{n}\'ak\inst{1}
}

\authorrunning{D. Kyselica et al.}
% \authorrunning{Anonymous}

\institute{
Zaitra s.r.o., Brno, Czech republic \\
\email{\{first\_name.last\_name\}@zaitra.io}
\and
Faculty of Informatics, Masaryk University, Brno, Czech republic 
}

\maketitle              % typeset the header of the contribution
% \institute{
% Zaitra s.r.o., Brno, Czech republic
% \email{\{first_name.last_name\}@zaitra.io}\\
% \url{http://www.springer.com/gp/computer-science/lncs} \and
% ABC Institute, Rupert-Karls-University Heidelberg, Heidelberg, Germany\\
% \email{\{abc,lncs\}@uni-heidelberg.de}}
%
\begin{abstract}
Natural disaster monitoring through continuous satellite observation requires processing multi-temporal data under strict operational constraints. This paper addresses flood detection, a critical application for hazard management, by developing an onboard change detection system that operates within the memory and computational limits of small satellites. We propose History Injection mechanism for Transformer models (HiT), that maintains historical context from previous observations while reducing data storage by over 99\% of original image size. Moreover, testing on the STTORM-CD-Floods dataset confirms that the HiT mechanism within the PrithVi-tiny foundation model maintains detection accuracy compared to the bi-temporal baseline. The proposed HiT-PrithVi model achieved 43 FPS on Jetson Orin Nano, a representative onboard hardware used in nanosats.
This work establishes a practical framework for satellite-based continuous monitoring of natural disasters, supporting real-time hazard assessment without dependency on ground-based processing infrastructure. 
Architecture as well as model checkpoints is available at 
% (*not present to preserve anonymity)
\url{https://github.com/zaitra/HiT-change-detection}


\end{abstract}

\section{Introduction}

According to the United Nations, between 2000 and 2019, 7,348 recorded natural disasters have affected 4.2 billion people with 1.23 million casualties~\cite{UNDissasters}. This represents a swift rise compared to the prior 20-year observation period, which documented 4,212 disasters. An early warning system can significantly mitigate these losses. Earth observations (EO) are particularly valuable in this case; e.g.  
optical sensors can detect almost every type of disaster, with the exception of earthquakes~\cite{denis2016evolution}.
With the rise of computer vision 
and the abundance of available data for training, thanks to Sentinel 1 \& 2 missions~\cite{sentinel}, the era of deep learning for EO has begun.
One of the earlier works builds upon AlexNet~\cite{krizhevsky2012imagenet}, and uses a similar Convolutional neural network for a classification problem for pre-disaster and post-disaster images of landslides and floods~\cite{amit2017disaster}. 

Nowadays, in the machine learning community, the problem of disaster monitoring falls under the change detection in general and is usually solved using semantic segmentation networks like UNet~\cite{ronneberger2015u} and Fully Convolutional Networks~\cite{long2015fully}. These models have been used for 
floods~\cite{nemni2020fully}, wildfires~\cite{rashkovetsky2021wildfire}, and landslides~\cite{bragagnolo2021convolutional} monitoring.

However, transmitting images on Earth for processing is not an ideal solution if our goal is to achieve the fastest possible response time. 
One way is to move processing from the ground to the satellite ~\cite{serief2023deep}, which poses challenges with limited computational and storage resources. 
Multiple studies have designed small neural networks for change detection systems, to meet onboard requirements~\cite{inzerillo2025compress}~\cite{ruuvzivcka2022ravaen}~\cite{herec2025sttorm}. 
Yet, still a bottleneck in these studies is the storage for historical observations used in change detection and their inability to efficiently store historical information from multiple time steps.
Moreover, these solutions do not account for the problem of continuous change detection.

With the arrival of devices designed for neural network inference on edge devices, such as Nvidia Jetson~\cite{jetson}, new opportunities have opened up to use more powerful foundation models. 
An example is Nvidia Jetson~\cite{jetson}, as its use onboard satellites has already been demonstrated, as in the case of the Forest-2 mission ~\cite{schottl2024real}

Foundation models, such as PrithVi~\cite{szwarcman2024PrithVi} or TerraMind~\cite{jakubik2025terramind}, also come in \textit{tiny} version with approximately 5 million parameters, allowing their use in this context and to take advantage of their pretraining on millions of images.

\paragraph{Contributions.}
To summarize, this work addresses the challenge of continual multi-temporal change detection for natural disasters using resource-constrained onboard devices. We propose a novel approach that leverages a compact History Embedding integrated into a ViT (Visual Transformer) based encoder, enabling efficient storage and processing of past observations. The following contributions highlight the key innovations and findings of our study:

\begin{enumerate}
    \item \textbf{HiT mechanism — a history injection transformer mechanism.} We introduce \textit{HiT}, a lightweight mechanism for injecting and updating a compact History Embedding inside a selected ViT block, enabling continual multi-temporal change detection without storing past images.
    
    \item \textbf{HiT-PrithVi — the first continual-change variant of PrithVi for onboard EO inference.} By integrating the HiT block into the distilled PrithVi-EO-2.0-tiny encoder and pairing it with an FPN decoder, we develop \textit{HiT-PrithVi}, a resource-efficient model suitable for devices such as Jetson Nano/Orin.
    
    \item \textbf{Reduction of onboard storage requirements.} We demonstrate that History Embeddings compress past observations by up to 99.6\% relative to raw Sentinel-2 tiles, allowing continental-scale historical memory to fit within a few gigabytes—making continuous monitoring feasible for small satellites.
    
    \item \textbf{Improved robustness to low-quality or missing temporal frames.} Experiments indicate that HiT-PrithVi maintains performance even when intermediate pre-disaster images are degraded, suggesting that the model successfully stores and reuses temporal information.
    
\end{enumerate}

\section{Related Work}

Change detection in remote sensing is a widely studied area and despite the disaster monitoring (namely wildfires, floods, landslides, etc.) also focuses on other tasks such as urban change or cover change detection. Change detection can be loosely defined as the difference between consecutive observations of the same area. 

% Siamese networks
In the case of two input images, the task is referred to as bi-temporal change detection. 
Most current solutions use modifications of Siamese networks
for change detections introduced in~\cite{daudt2018fully}. The main idea is the use of shared encoder for both input images in a UNet-like\cite{unet} convolutional network. The authors in~\cite{bandara2022transformer} used this idea and used a transformer\cite{vaswani2017attention} based encoder, named ChangeFormer, and computed differences of the images at multiple scales before upsampling into the binary change mask. This idea was extended in~\cite{xu2025pushing}, where the authors enriched the difference information at various depths using a custom Local-Global Fusion Block based on a self and global attention mechanisms. 
Taking advantage of the Segment Anything (SAM) family of models ~\cite{kirillov2023segment}, the authors~\cite{ding2024adapting} used a frozen FastSAM~\cite{zhao2023fast} encoder in an Unet-like structure with learnable adapters. 
A different approach was developed in~\cite{kim2021graph} using Graph Neural Networks~\cite{kim2021graph}. Firstly, for each input image, a graph is constructed using Faster RCNN for object detection, with edges computed as relative Euclidean distances between objects in the image. The resulting graphs are input to the GNN with a binary classification head. 
In real use, these scenarios require the storage of the latest image on the device, which can be a limiting factor if monitoring a large area is required.

% On board systems 
To tackle issues with limited storage and computation power, ~\cite{serief2023deep} decided to choose a slightly different approach, using unsupervised machine learning. The input images are still preprocessed by the shared CNN encoder, and the change regions are identified using the k-means clustering algorithm~\cite{kmeans} in the feature differences to group the changes and no changes to the pixels in distinct classes. Another branch of solutions uses Variational Auto-Encoders (VAE)~\cite{kingma2013auto} to compress the input images into compact embeddings with fractional memory requirements. Moreover, the change detection can then be performed in compressed form. 
The first method of this kind for onboard change detection is the RaVAEn model~\cite{ruuvzivcka2022ravaen}. The VAE model was trained in an unsupervised manner to encode $32\times32$ tiles. The cosine difference between two encoded representations determines the magnitude of the change in the area. This work inspired the development of the STTORM-CD~\cite{herec2025sttorm} model, which has improved the original method by incorporating labeled data into the training process and the use triplet loss.
A very similar approach was demonstrated in~\cite{yadav2024unsupervised}, where the authors also used contrastive learning techniques with the addition of LSTM~\cite{hochreiter1997long} and 3D Convolution blocks for better temporal and spatial feature extraction. 
Still, training such small models from scratch does not always lead to optimal performance.

% Knowledge distillation
A complementary strategy is to use larger and better-performing models to train a smaller version, which can be on edge-devices in a process called knowledge distillation~\cite{hinton2015distilling}. Experiments with model compression limits have been conducted in~\cite{pang2024exploring}, where the authors trained the target small model, for object recognition, in a pyramidal structure by incrementally decreasing the size of the student and teacher models at each stage. Moreover, specifically for drought monitoring, the authors~\cite{zhang2024domain} used knowledge distillation to train a smaller version of their Variational recurrent network, producing a smaller model with better performance as if it were trained separately.  

% Foundation models
In recent years, the increased popularity and successes of foundational models in other areas, such as natural language processing, have motivated their adaptation to the EO domain. 
Thanks to the availability of large datasets mainly using data from Sentinel 2~\cite{sentinel} and Landsat~\cite{williams2006landsat} missions, training billion-parameter models in a self-supervised manner becomes possible. This led to the development of multiple datasets and models including Satlas\cite{bastani2023satlaspretrain} (SWIN~\cite{liu2021swin}, ResNet~\cite{he2016deep}), SSL4EO~\cite{wang2023ssl4eo}(ViT~\cite{dosovitskiy2020image}, ResNet), family of PritVi models\cite{szwarcman2024PrithVi} or TerraMind models\cite{jakubik2025terramind}. The last two mentioned also offer smaller distilled versions below the 5 million parameter, which can be used on some edge devices with sufficient resources, such as Jetson Orin~\cite{jetson}. 

% Continual Change detection
However, research on continual change detection for onboard processing is extremely limited, practically nonexistent.
% Nevertheless, the available research in the area of continual change detection is very limited in a special case of onboard processing, practically nonexistent. 
One study explored continuous change detection in an urban context\cite{hafner2025continuous}. The method first used a shared encoder for all input images producing feature maps at different stages, which are then processed by a temporal feature refinement blocks with attention heads. The refined feature maps, together with their differences in time, are sent to the decoder to produce segmentation and change detection outputs. 
Although this method is not suitable for direct on board use, as it requires access to all historical images during inference, it represents a step in the right direction in solving continuous change detection problem.

\section{Method}

\subsection{Model}

\begin{figure}[h!]
    \centering
    \includegraphics[width=\linewidth]{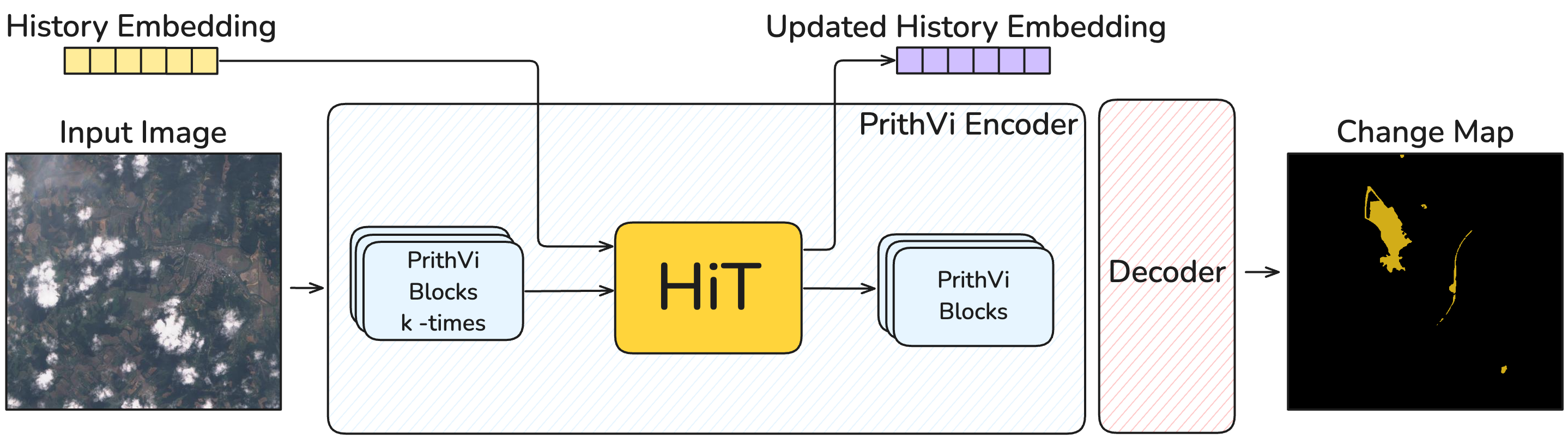}
    \caption{Overview of the proposed method with History Injection mechanism (HiT) inside PrithVi transformer encoder. HiT uses historical information stored in History Embedding to inform the change detection model.}
    \label{fig:method}
\end{figure}

Our proposed model has an encoder-decoder structure with some modifications.
As our target device is Jetson Nano, we decided to take advantage of the pretrained PrithVi-EO-2.0-tiny\cite{szwarcman2024PrithVi} model as the encoder with History injection mechanism to incorporate the History Embedding (HE) in the input (see Sec.~\ref{sec:embed}). As a decoder architecture we chose Feature Pyramidal Network for Semantic Segmentation (FPN)~\cite{kirillov2017unified}. The implementation of the models with their pretrained weights was obtained from the TerraTorch~\cite{gomes2025terratorch} and Segmentation Models Pytorch~\cite{Iakubovskii:2019} Python libraries. The high-level view of the final model is presented in Fig.~\ref{fig:method}.

\newpage

\subsection{History Injection Mechanism for Transformers: HiT}\label{sec:embed}

\begin{figure}
    \centering
    \includegraphics[width=0.8\linewidth]{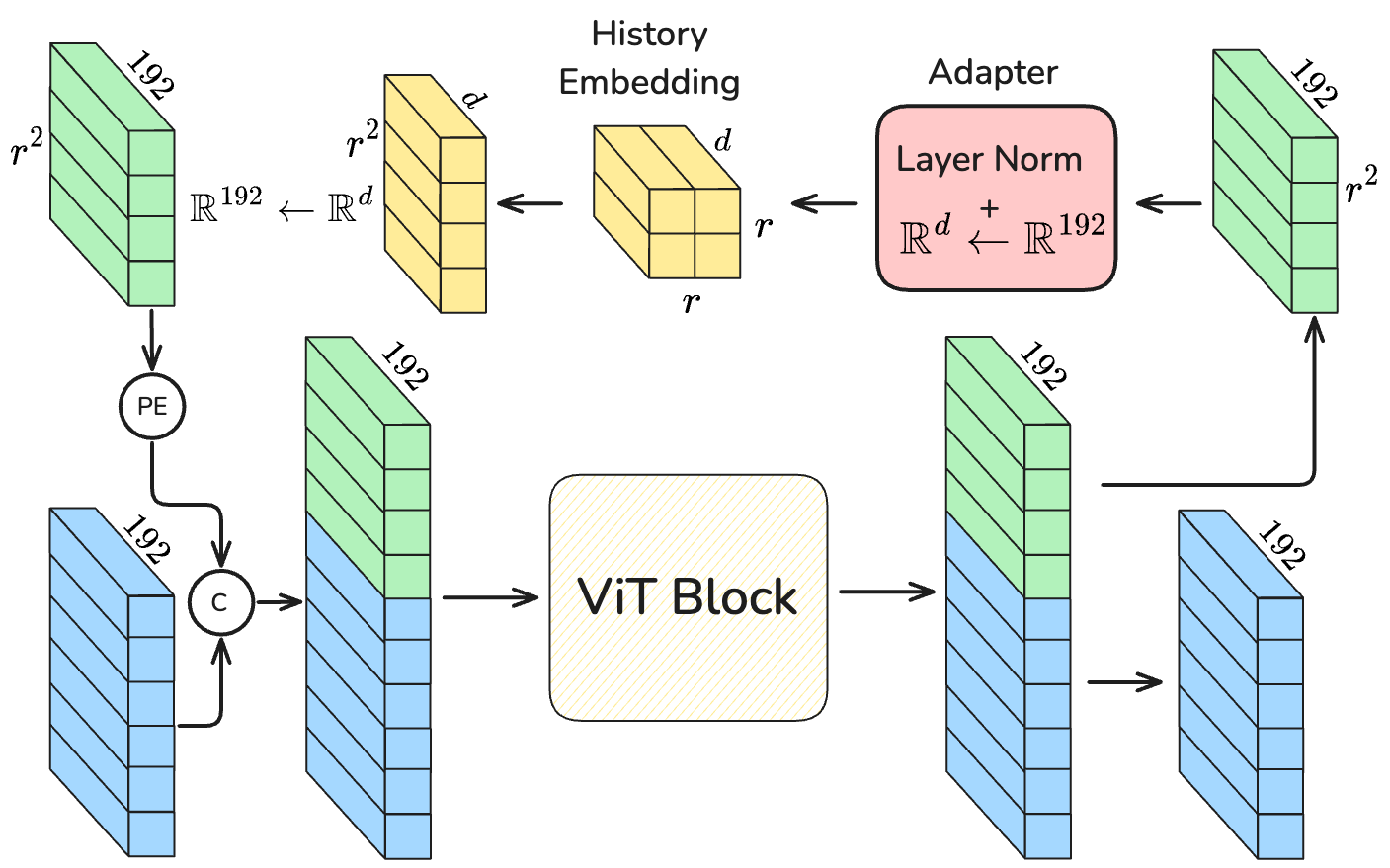}
    \caption{HiT mechanism. The History Embedding is transformed to the required dimensionality and fed into the ViT block. The new embedding is extracted from the block's output. Image input tokens are shown as blue, embedding in green (projected version) and yellow. (PE - position encoding, C - concatenation)}
    \label{fig:embed}
\end{figure}

The ViT model processes an input image as a sequence of $16x16$px patches. Each patch is projected into a 1-D vector of size 192 called a token. During this process, positional encoding is added to the token to store its spatial location of the original patch. These tokens then pass through multiple ViT blocks, the number of which depends on the actual architecture, 
to gradually enrich the information present in the tokens.
We leverage the fact that ViT block performs cross attention between all input tokens, thus information from one particular input token will influence all output tokens. 
Due to its ability to handle an arbitrary number of input tokens, we can simply inject HE into any block. Therefore, the original input tokens will be enhanced by the historical information, and HE tokens will be updated with the recent information.
The updated HE is an output of ViT block at stage $k$, therefore it cannot be directly used as input for stage $k$. 
This problem is solved by introducing an adapter block, consisting of a layer norm plus a linear layer, approximating the inverse function of the ViT block.
Moreover, the linear layer inside the adapter can be used to reduce the dimensionality of HE for more efficient storage. Before processing the next input, the HE is decompressed by a linear layer to its original dimensionality.
The HiT mechanism is visualized in Fig.\ref{fig:embed}. \\

In this case, we used the PrithVi-EO-2.0 tiny ViT model with 11 blocks. Input images of size $256\times256$ are divided into $16\times 16 $ patches yielding $\frac{256}{16}. \frac{256}{16} = 256$ tokens.  
The History Embedding $E\in \mathbb{R}^{r\times r\times d}$ represents historical information for a particular area in a grid $r\times r$. The parameters $r$ and $d$ both control the amount of stored information, but negatively influence the memory footprint of the system. 
$E$ is concatenated with the input tokens of the $k^{th}$ PrithVi block, but it is first flattened and projected by a linear layer to shape $r^2\times 192$, to match the dimensions of the ViT token, and positional encodings are applied. The ViT block then outputs $r^2+256$ tokens, which are subsequently separated back to the History Embedding and processing tokens. HiT mechanism ensures that the historical information in the embedding is incorporated into the processing of the current image. Likewise, the information stored in the  History Embedding is updated with the current image through the nature of the ViT block.
The last step is to project $E$ to the original dimension and to apply the inverse ViT block function via the adapter block.

In summary, the HiT mechanism is parametrized by injection depth $k$, HE grid size $r$,and HE dimension $d$.

\section{Data}

Unfortunately, no existing dataset can be used directly to train and test models for continuous multi-temporal disaster monitoring. 
As a substitute, we have chosen two multi-temporal Sentinel 2 flood datasets:
RaVAEn-Floods~\cite{ruuvzivcka2022ravaen} and STTORM-CD-Floods~\cite{herec2025sttorm}.
Importantly, the continuous nature will be added artificially to the training process via CutMix in the augmentation process (see Sec.~\ref{sec:preprocessing}).  
The choice of floods over other disaster types was made because of the fact that the change detection tends to be more difficult, as it often contains detailed changes and also depends heavily on the prior knowledge of the area. In contrast to wildfires, which often spread over large areas and are easily distinguishable in infrared bands. The RaVAEn-Floods dataset contains flood scenes besides other disasters, but not in sufficient numbers for training. Furthermore, there are no predefined test and training splits, as it was designed. Therefore, STTORM-CD-Floods was added to the training process. Moreover, its predefined and hand-picked test set will serve as a robust test set for model evaluation.
The final sample counts are presented in Tab.~\ref{tab:dataset1}.

\setlength{\tabcolsep}{3pt} % Default is usually 6pt
\begin{table}[ht]
    \centering
    \caption{Composition of the train and test set.}
    \label{tab:dataset1}
    
    \begin{tabular}{c|c|c|c}
\toprule

\multirow{2}{*}{\textbf{Split }} & \multirow{2}{*}{\textbf{Dataset}} & \multirow{2}{*}{\textbf{ Number of Events }} & \textbf{ Number of Tiles }  \\
& & & ($256\times 256$ px) \\
\midrule
\multirow{2}{*}{\textbf{Train }} & \text{ RaVAEn-Floods}~\cite{ruuvzivcka2022ravaen} & 4 & 219 \\
  & \text{ STTORM-CD-Floods}~\cite{herec2025sttorm}\text{ } & 12 & 372  \\
\midrule
\textbf{Test} & STTORM-CD-Floods~\cite{herec2025sttorm} & 4 & 134  \\
\bottomrule
    \end{tabular}
\end{table}

\subsection{Preprocessing \& Augmentation}\label{sec:preprocessing}
For model training and testing, basic preprocessing and augmentation techniques were employed. First, the images were cut into individual tiles of $256\times 256$ pixels. A target segmentation map was generated between each pair of images in the input sequence.

The original data from Sentinel 2 contains all 13 bands, but since PrithVi was pretrained only on 6 of them, we used the same selection of bands: B02, B03, B04, B08, B11 and B12. Min-max normalization was used with values 0 and 10,000.

Additionally, random horizontal/vertical flip, color jitter, and random rotation augmentations were applied to the training data. 
However, the data does not naturally contain continuous changes. Therefore, a modified CutMix~\cite{yun2019cutmix} 
augmentation was applied to simulate continuous changes, where change events were randomly inserted at different positions in the temporal sequence, forcing the model to learn from continuous change scenarios.

\begin{figure}
    \centering
    \includegraphics[width=0.9\linewidth]{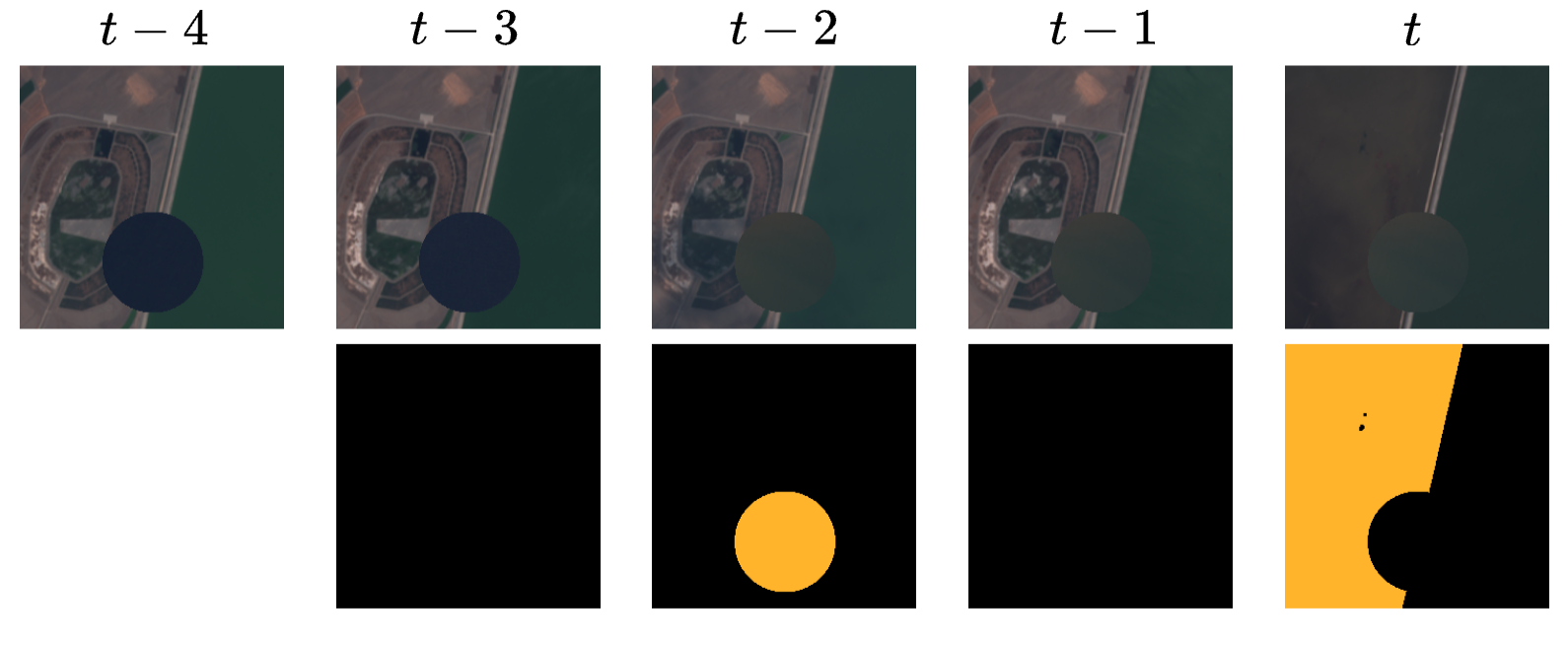}
    \caption{Result of modified CutMix augmentation. Top line shows image sequence, bottom line shows corresponding change mask. A change (in orange) is introduced in step $t-2$.}
    \label{fig:cutmix}
\end{figure}

\section{Experiments}

\subsection{Training Setup}

For reproducibility and mitigation of random factors, for each set of parameters, we trained 3 models with random initial parameters for 50 epochs on a NVIDIA GeForce RTX 3090.
Afterwards, the checkpoints with the highest F1-score for the change class were selected and used for comparison. The optimization was performed using AdamW optimizer~\cite{loshchilov2017decoupled} with learning rate $10^{-4}$ and Cosine Annealing scheduler~\cite{loshchilov2016sgdr}. The basic model configuration contains the HiT block in stage 5 of the PrithVi encoder. The HE consists of $16\times 16=256$ tokens of size $192$.

The loss is computed as the sum of the Cross entropy~\cite{mao2023cross} and the Dice loss~\cite{sudre2017generalised} for each image in the series. Additionally, the loss is computed for the case when an after-disaster image is introduced after each before-disaster image in the series. 
For reference, a baseline bi-temporal model, with the PrithVi encoder and FPNDecoder, is trained in the same manner.

% \subsection{Experiment I: Injection Depth}
\subsection{Injection Depth}

\begin{figure}[ht]
    \centering
    \includegraphics[width=.75\linewidth]{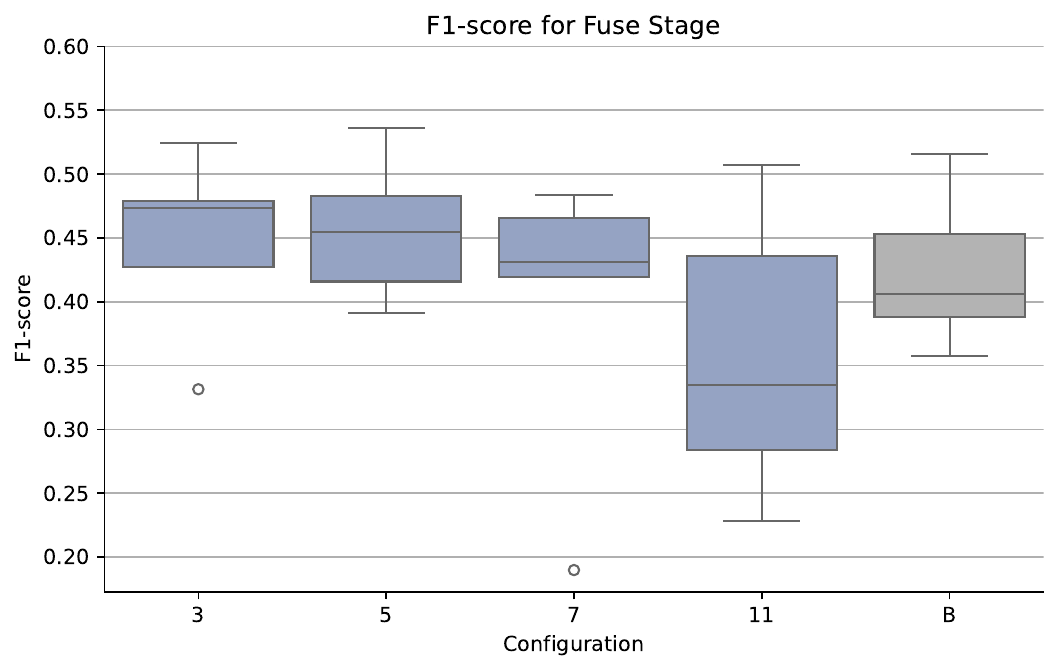}
    \caption{Model performance with varying History Embedding fusion stage and the bi-temporal baseline B.}
    \label{fig:fuse_stage}
\end{figure}

Encoders iteratively extract important features from the input. Feature maps created by earlier ViT blocks produce more low-level features, whereas later blocks produce high-level features. HiT inserts HE at injection depth $k$ , meaning in the $k^{th}$ block.  
We tested four different fusion stages for HiT: 3,5,7, and 11 since PrithVi-tiny contains 11 blocks. The results presented in Fig.~\ref{fig:fuse_stage}, show that the earlier fusion stages performed slightly better. This may be caused by the fact that 
floods usually exhibit fine-grained differences that are more visible
in low-level features. Moreover, the performance of the model is comparable to the bi-temporal baseline marked B.

% \subsection{Experiment II: Dimensionality Reduction}
\subsection{HE Dimensionality Reduction}

\begin{figure}[ht]
    \centering
    \includegraphics[width=.75\linewidth]{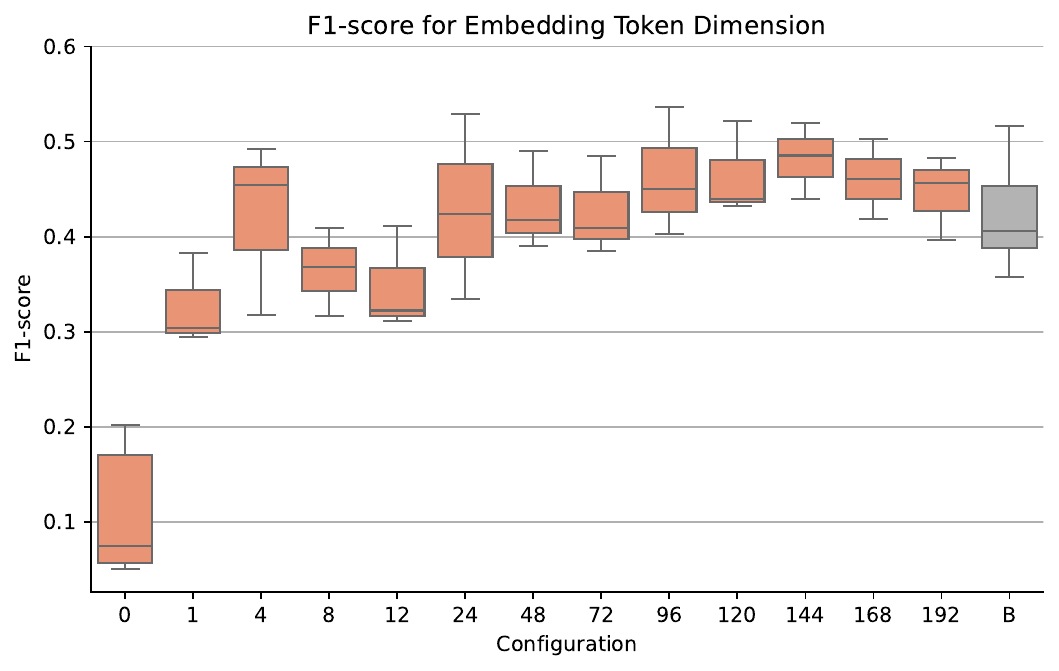}
    \caption{Model performance with varying History Embedding dimension and the bi-temporal baseline B.}
    \label{fig:dim}
\end{figure}

The second free parameter for HiT is 
the size/dimension of individual tokens in the HE, directly controlling the compression rate of HE. Therefore, it influences both the storage requirements of the target device and the expressive power of the model.  
The PrithVi default is $192$.
We tested 13 possible values to identify any clear relationship
between the embedding token dimension and F1-Score. As shown in Fig.~\ref{fig:dim}, there is a notable drop in performance for small values $< 24$. Surprisingly, the model with values $\geq 24 $, achieved
a performance similar to both the baseline B and the PrithVi default dimension of 192. 
Another observation is that with lower values of token size, the variance of the results is higher. 

% \subsection{Experiment III: Spatial Reduction}
\subsection{HE Spatial Reduction}

\begin{figure}[ht]
    \centering
    \includegraphics[width=.75\linewidth]{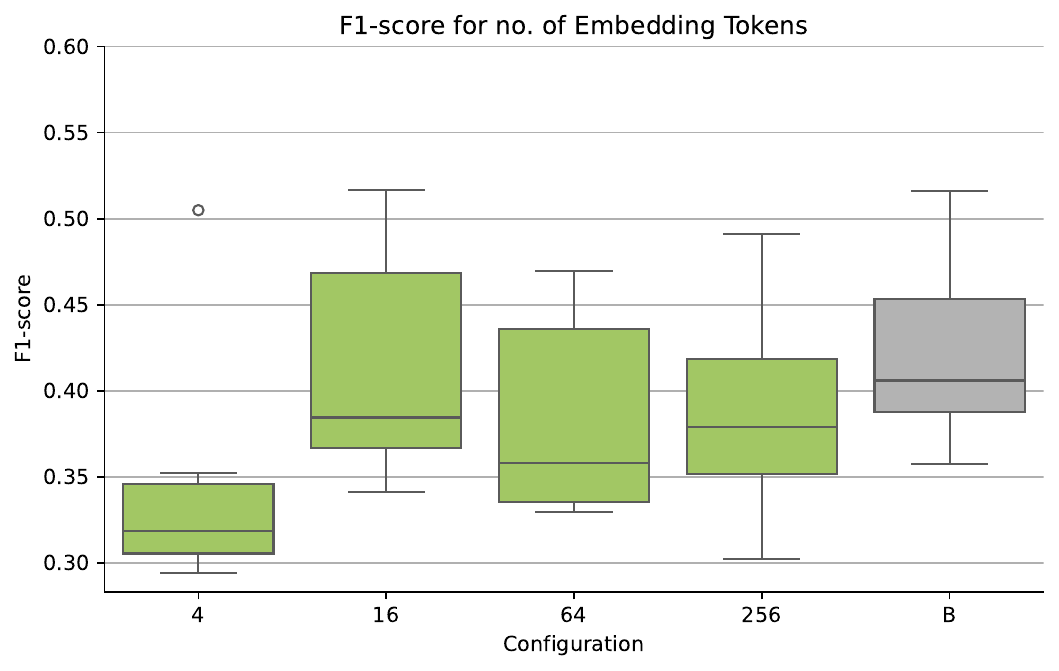}
    \caption{Model performance with varying History Embedding spatial dimension and the bi-temporal baseline B.}
    \label{fig:tokens}
\end{figure}

Another way to reduce the storage requirements of the HE is to reduce its size in spatial dimensions. Therefore, we tested four different sizes of its grid, thus simulating a pooling operation. Grid sizes 2, 4, 8, and 16 result in 4, 16, 64, and 256 tokens, respectively. The experimental results in Fig.~\ref{fig:tokens} show that the number of tokens 
strongly influenced both the mean performance and its variance
Still, as few as 16 tokens are sufficient for the model to learn representations with only a slight drop in performance compared to the baseline.

% \section{Results}
\section{Results and Discussion}
\label{sec:results}

% In this section, we evaluate the performance of the proposed History-Injection Transformer (HiT) against several baselines, analyze its architectural stability, and discuss its practical implications for onboard deployment.

% \subsection{Quantitative Performance and Stability}
In this section, we evaluate the performance of the HiT-PrithVi ($d=24$, $r^2=64$, $k=5$) against against the bi-temporal PrithVi-tiny baseline and ContUrbanCD~\cite{hafner2025continuous}, a model originally designed for urban change detection. To ensure statistical significance, all models were trained five times from different random initializations. The results are summarized in Table~\ref{tab:comparison}.

\begin{table}[h]
\small
\centering
\caption{Model performance comparison across five independent runs.}
\label{tab:comparison}
\begin{tabular}{@{}l|c|c|c|c|cc@{}}
\toprule
\textbf{Model} &  \textbf{F1} & \textbf{Precission} & \textbf{Recall} &  \textbf{Parameters} & \textbf{Input Size} \\ \midrule
% \multirow{2}{*}{\textbf{Model}} & \multicolumn{2}{c}{\textbf{F1}} & \textbf{Precision}  &\textbf{Recall}& \textbf{Params.} & \textbf{Input Size}  \\ 
% & \textbf{Best} &\textbf{Mean} & \\ \midrule
Baseline  &  0.41 $\pm$ 0.06 & 0.73 $\pm$ 0.05 &0.29 $\pm$ 0.05 & $8.5$M & $2\times$  \\
\textbf{HiT-PrithVi} & 0.38 $\pm$ 0.08 & 0.70 $\pm$ 0.03&0.27 $\pm$ 0.08 & \textbf{7.8M}  & $\mathbf{1.004\times}$  \\
ContUrbanCD  & 0.46 $\pm$ 0.26 & 0.82 $\pm$ 0.06 &0.35 $\pm$ 0.25 & $25$M & $n \times$ & \\ \bottomrule
\end{tabular}
\end{table}

The results highlight a critical trade-off between peak performance and operational reliability. While ContUrbanCD achieved the highest single-run F1-score $0.740$ (compared to HiT-Prithvi with $0.48$ and bi-temporal baseline with $0.52$), its high standard deviation $\pm0.26$ indicates extreme training instability and a tendency to overfit. Furthermore, with 25M parameters, ContUrbanCD exceeds the memory constraints of our target hardware, the Jetson Orin Nano (8GB), and is included here only as a performance reference.
In contrast, \textbf{HiT} achieves a more favorable balance for onboard deployment. Despite being smaller than the baseline (7.8M vs 8.5M parameters), it maintains comparable mean performance. Most importantly, HiT drastically reduces the required input data size from $2\times$ (two full images) to nearly $1\times$ ($1.0039\times$) by utilizing History Embedding (see Section~\ref{sec:footprint}). This reduction is pivotal for continuous monitoring, as it minimizes the onboard memory overhead required to store historical reference states.

\begin{figure}
    \centering
    \includegraphics[width=.97\linewidth]{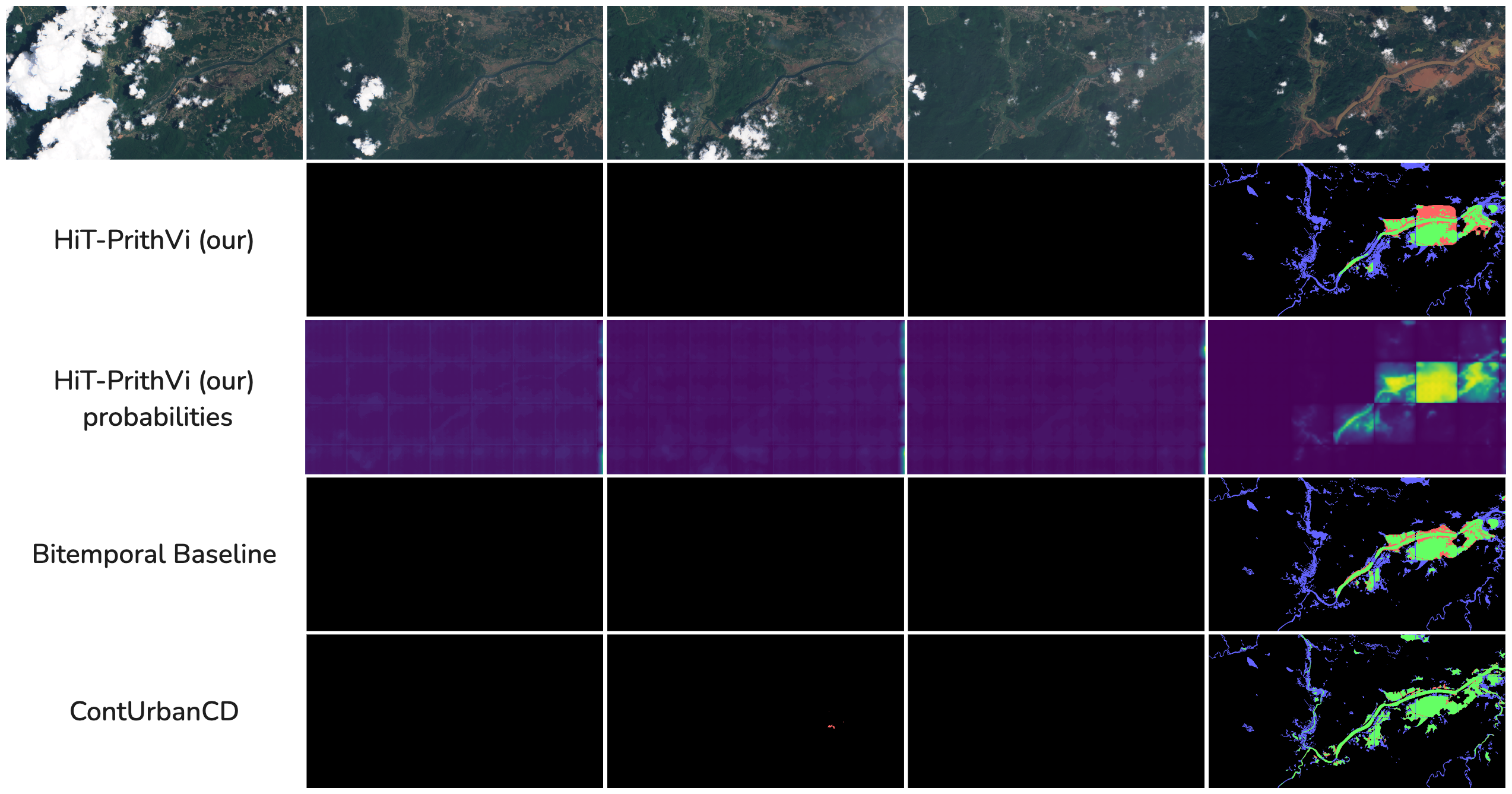}
    \caption{Model's results for the Laos test scene. Rows from top to bottom: input series, HiT-PrithVi prediction, HiT-PrithVi probabilities, bi-temporal baseline predictions, ContUrbanCD predictions. Segmentation results: true positive (green), false positive (red), false negative (blue), true negative (black).}
    \label{fig:laos-results}
\end{figure}

For visual reference, the output of the models for the Laos test scene are presented in Fig.~\ref{fig:laos-results}. It is important to emphasize that HiT-PrithVi and baseline model were not able to successfully detect small changes, probably due to number of training parameter size and amount of training data. 
% (complete test results are in Appendix~\ref{app:results}). 
% However, this is not an issue in real-world applications, as in that case the image can be sent to Earth for further processing.
Surprisingly, for images without change, the model produces almost no false positives, which is really important for the target use case. We discuss the errors in Section~\ref{sec:error_analysis}. 

\subsection{Resource Efficiency: Memory footprint}\label{sec:footprint}

When comparing the memory footprint of the HE ($d=24$, $r^2=64$) with the original image (1.57MB), it saves $99.61\%$ of memory. Therefore, for monitoring the area of Europe without Russia $4.85$ GB is needed, and $139.63$ GB for the World land area. Such storage requirements are feasible on board small satellites. Table~\ref{tab:memory} shows results for a wide range of possible settings.

\begin{table}[ht]
    \centering
    \caption{Memory footprint for different configurations of History Embedding.}
    \label{tab:memory}
    
    \begin{tabular}{c| c |r|r|r}
\toprule
\textbf{ Size } & \textbf{ \# Tokens } & \textbf{ \% of Image }  & \textbf{  Europe  } & \textbf{  World  } 
\\
\midrule
 \multirow{4}{*}{192}       & 256        & 12.5  \%   & 155.32  GB & 4.47 TB     \\
        & 64         & 3.125 \%   & 38.83  GB  & 1.12 TB \\
        & 16         & 0.781 \%   & 9.71  GB   & 279.26  GB  \\
        & 4          & 0.195 \%   & 2.43 GB    & 69.82 GB    \\
 \midrule
 168       & \multirow{5}{*}{256}        & 10.938\%   & 135.91 GB  & 3.91 TB     \\
 % 168       & \multirow{9}{*}{256}        & 10.938\%   & 135.91 GB  & 3.91 TB     \\
 % 144       &         & 9.375 \%   & 116.49 GB  & 3.35 TB     \\
 120       &         & 7.813 \%   & 97.08  GB  & 2.79  TB    \\
 % 96        &         & 6.25  \%   & 77.66  GB  & 2.23  TB    \\
 72        &         & 4.688 \%   & 58.25  GB  & 1.68 TB     \\
 % 48        &         & 3.125 \%   & 38.83  GB  & 1.12  TB    \\
 24        &         & 1.563 \%   & 19.42  GB  & 558.53 GB   \\
 % 12        &         & 0.781 \%   & 9.71  GB   & 279.26  GB  \\
 8         &         & 0.521 \%   & 6.47   GB  & 186.18 GB   \\
 \midrule
 \textbf{24}        & \textbf{64}         & \textbf{0.391 \%}  & \textbf{4.85  GB}   & \textbf{139.63 GB}   \\
\bottomrule
    \end{tabular}
\end{table}

\newpage
\subsection{Temporal Persistence and Prediction Stability}

% the following experiment,to observe

\begin{figure}
\centering
\begin{subfigure}{\textwidth}
    \centering
    \includegraphics[width=.97\textwidth]{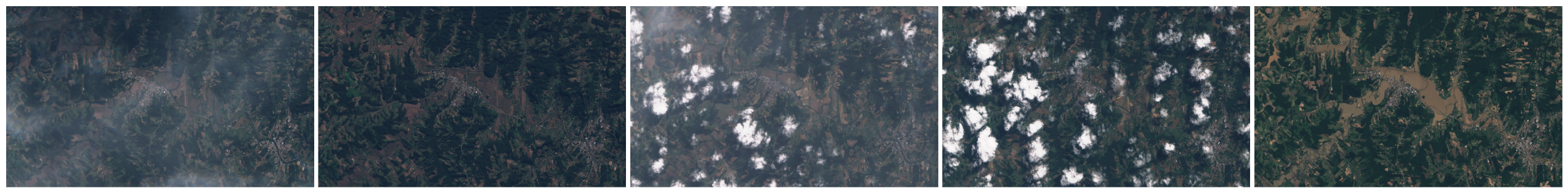}
    % \caption{}
    % \label{fig:brazil-input}
\end{subfigure}
\begin{subfigure}{\textwidth}
    \centering
    \includegraphics[width=.97\textwidth]{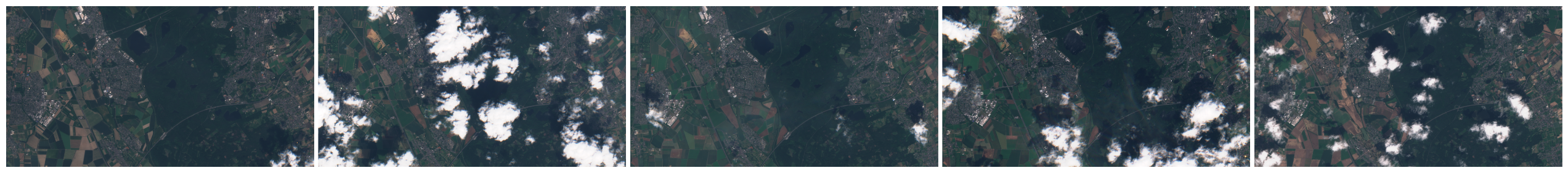}
    % \caption{}
    % \label{fig:germany-input}
\end{subfigure}
\caption{Test scenes from STTORM-CD-Floods~\cite{herec2025sttorm} (a) Brazil, (b) Germany.}
\label{fig:brazil-germany-inputs}
\end{figure}

To evaluate the model's capacity to continuously store historical information, we designed an experiment to observe how it leverages information from the HE during the processing of low-quality inputs.
To isolate these low-quality images, we employed a bi-temporal baseline model. In this setup, each pre-disaster image in a series was individually paired with the post-disaster image for inference. The underlying logic is straightforward: if the baseline model yields a substantially lower performance on a specific pair compared to the rest of the series, that particular pre-disaster image is deemed to be of low quality. By analyzing the drop in F1-score between the best overall result and each individual pre-disaster image, we successfully identified two primary low-quality candidates: Brazil No. 3 and Germany No. 3. The comprehensive analysis can be found in Table~\ref{tab:time} of Appendix~\ref{app:bad_images}, with the corresponding test scenes depicted in Fig.~\ref{fig:brazil-germany-inputs}.

\begin{table}[ht]
    \centering
    \caption{HiT-PrithVi model F1-Score for pre-disaster sub-series.}
    \label{tab:final-time}
    
    \begin{tabular}{c|c|c|c|c}
\toprule
\multirow{2}{*}{\textbf{Scene}} & \multicolumn{4}{c}{\textbf{Sub-series size}} \\
 & \textbf{ 1 } & \textbf{2} & \textbf{3} & \textbf{4} \\
\midrule
Brazil  & 0.4458 & 0.5321 & \textbf{0.5335} & 0.5112 \\
Germany & 0.0444 & 0.4578 & \textbf{0.4605} & 0.4550 \\
\bottomrule
\end{tabular}
\end{table}

Notably, while Brazil No. 1 produced the worst score, it could not be utilized for this test. Because it is the first image in the series, the HE is empty at that moment, precluding any measurement of continuous memory. 
Our HiT-PrithVi model was evaluated analogously. During series processing, it was prompted to create a prediction for the post-disaster image. As shown in Table~\ref{tab:final-time}, the HiT model did not experience any drop in performance when presented with low-quality images before the final prediction. This suggests its capability to store important information in the HE. However, this requires further investigation, which is not possible without specialized testing data.

\subsection{Error Analysis}\label{sec:error_analysis}

\begin{figure}[h]
    \centering
    \includegraphics[width=0.95\linewidth]{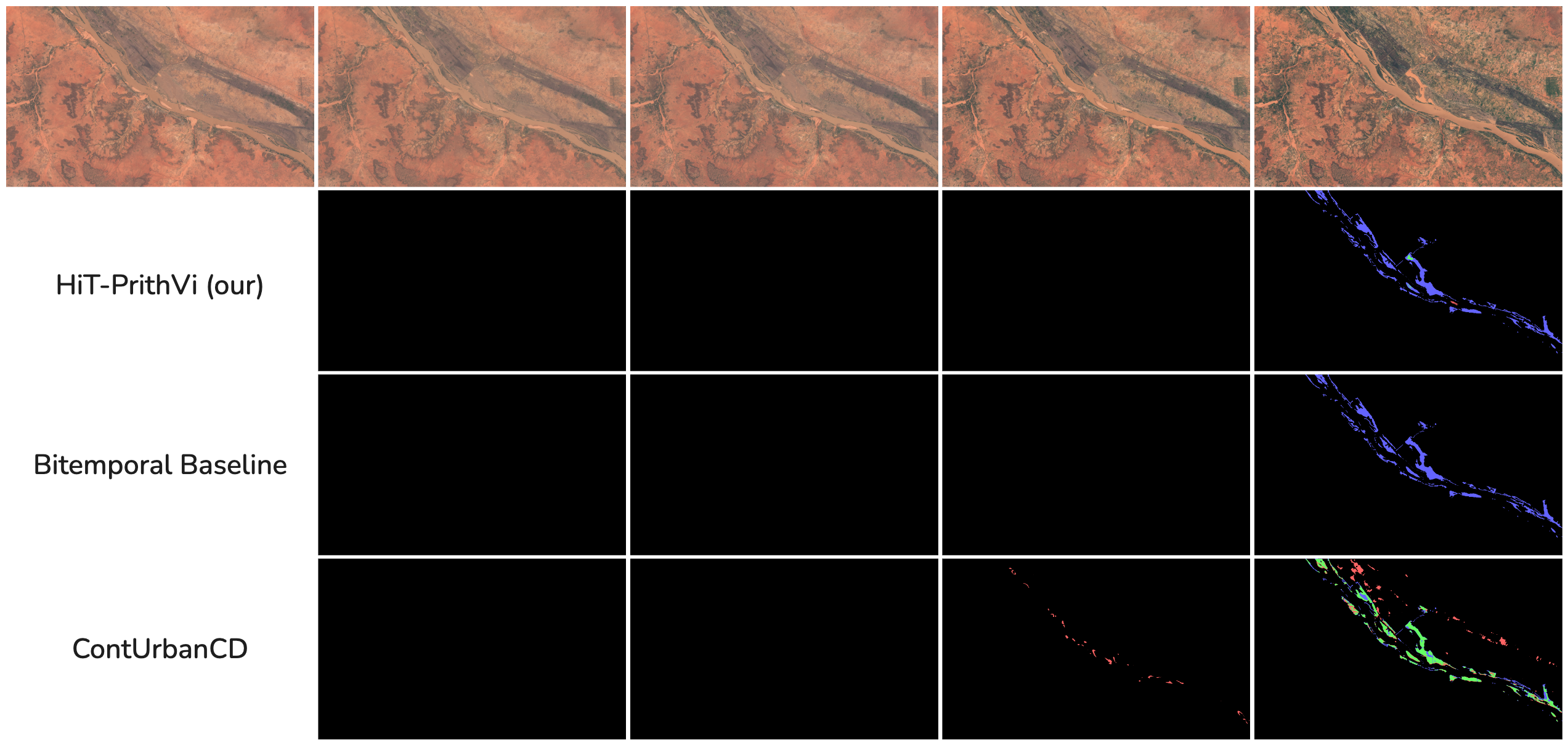}
    \caption{HiT-PrithVi, bi-temporal baseline, ContUrbanCD results for the Niger scene.Segmentation results: TP (green), FP (red), TN (blue), TN (black).}
    \label{fig:niger}
\end{figure}

\begin{figure}
    \centering
    \includegraphics[width=.95\linewidth]{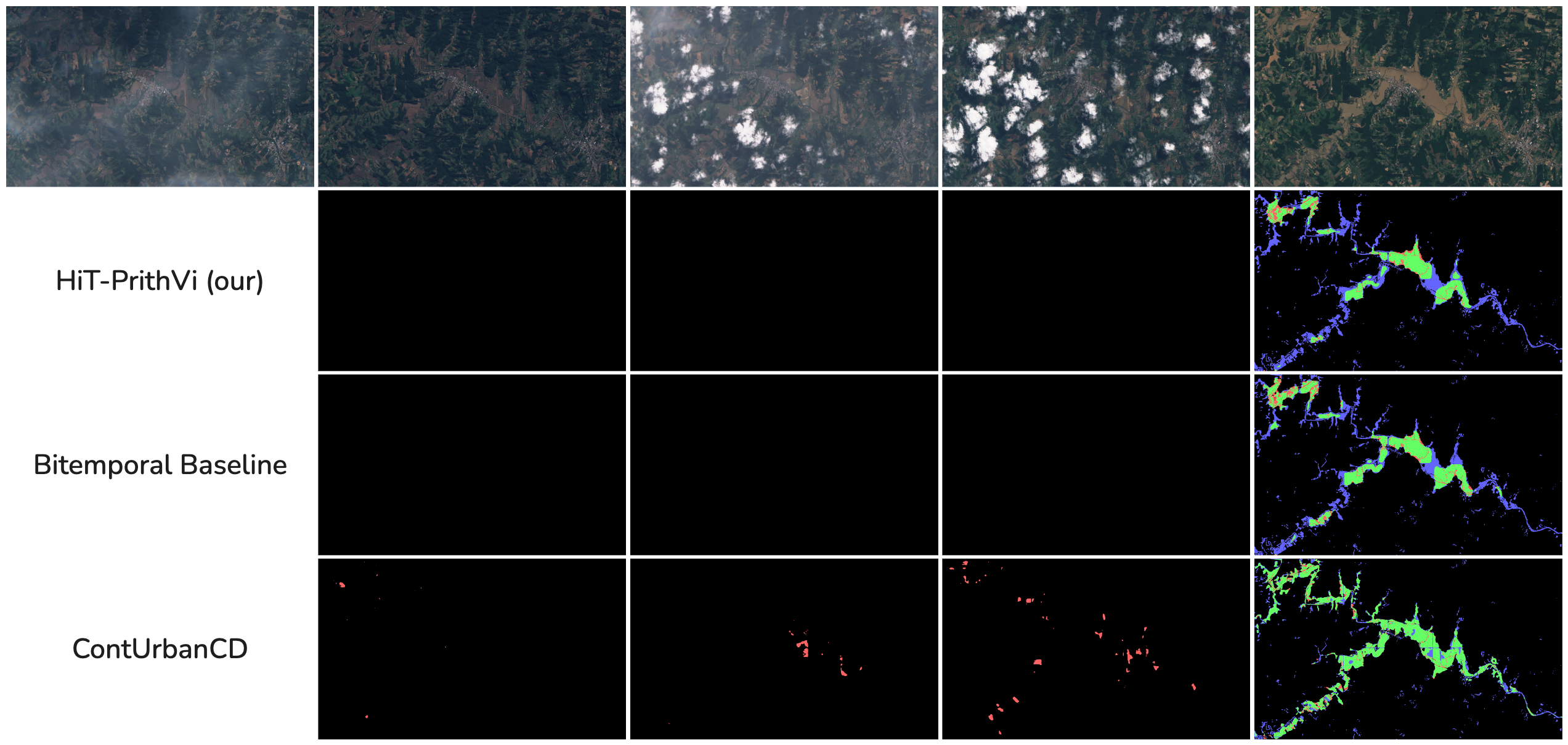}
    \caption{HiT-PrithVi, bi-temporal baseline, ContUrbanCD results for the Brazil scene. Segmentation results: TP (green), FP (red), TN (blue), TN (black).}
    % \caption{HiT-PrithVi, bi-temporal baseline, ContUrbanCD results for the Brazil scene.Segmentation results: true positive (green), false positive (red), false negative (blue), true negative (black).}
    \label{fig:brazil}
\end{figure}

The model failure can be seen on the Niger test scene from STTORM-CD-Floods in Fig.~\ref{fig:niger}. This scenario tests the model`s generalization on biome not present in the training set. HiT-Pritvi was able to identify only a small region with change, while the baseline model failed completely. The reference ContUrbanCD model detects change precisely with the cost of false positive detections in the timesteps with no change. This problem is even more prevalent on the Brazil scene in Fig.~\ref{fig:brazil}. Although both baseline and HiT-PrithVi have correctly predicted no change in first 3 predictions, ContUrbanCD has again produced false positive change predictions. For real life use cases, HiT-PrithVi behavior is preferable to ContUrbanCD, as false positive detections in no change image will trigger costly downlink of the scene.

% \begin{figure}
% \centering
% \begin{subfigure}{\textwidth}
%     \centering
%     \includegraphics[width=\textwidth]{images/niger.png}
%     \caption{}
%     \label{fig:niger}
% \end{subfigure}
% \begin{subfigure}{\textwidth}
%     \centering
%     \includegraphics[width=\textwidth]{images/brazil.png}
%     \caption{}
%     \label{fig:brazil}
% \end{subfigure}
        
% \caption{Model's results for the Niger (a) and Brazil (b) test scenes. Rows from top to bottom: input series, HiT-PrithVi prediction, bi-temporal baseline predictions, ContUrbanCD predictions. Segmentation results: true positive (green), false positive (red), false negative (blue), true negative (black)}
% \label{fig:figures}
% \end{figure}

\section{Conclusion}

This work addresses the challenge of continuous onboard change detection for natural disaster monitoring by introducing the History Injection mechanism for Transformers (HiT). Our proposed HiT-PrithVi model integrates compact History Embeddings directly into the PrithVi-EO-2.0-tiny encoder, enabling efficient multi-temporal flood detection under the strict constraints of small satellite platforms.
The HiT mechanism reduces storage requirements by 99.61\% compared to storing raw imagery, allowing continental-scale monitoring with only a few gigabytes of memory. Despite this dramatic compression, the model achieves an F1-score of 0.479 on the STTORM-CD-Floods test set—comparable to bi-temporal baselines—while producing almost no false positives on scenes without change. Real-time inference at 43 FPS on Jetson Orin Nano 8GB demonstrates the model's practical viability for onboard deployment.
Our experiments reveal that the model is robust to variations in fusion stage, embedding dimensionality, and spatial resolution, with optimal performance achieved using 64 tokens of dimension 24 and injection depth 5. Analysis on the STTORM-CD-Floods Brazil scene suggests that the History Embedding successfully retains temporal information, maintaining performance even when intermediate observations are degraded.

Future work should focus on developing specialized datasets for continuous change detection to enable a more thorough evaluation of temporal persistence and multi-event tracking capabilities. Nevertheless, this work establishes a practical framework for autonomous satellite-based disaster monitoring and demonstrates that foundation models can be adapted for resource-constrained edge computing scenarios. We hope that the HiT mechanism will inspire further research in efficient onboard Earth observation systems and extend to other disaster types and foundation model architectures.

\section*{Acknowledgment}

\noindent
\begin{minipage}{0.15\textwidth}
    \includegraphics[width=0.9\linewidth]{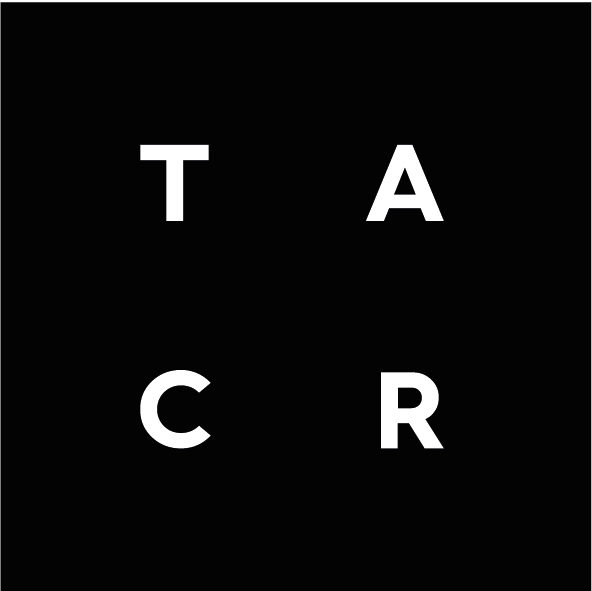}
\end{minipage}\hfill
\begin{minipage}{0.87\textwidth}
    This project is co-financed from the state budget by the
    Technology agency of the Czech Republic
    under the SIGMA Programme, Project no. TQ16000010. 
    Grammatical corrections and sentence rephrasing
    were done using ChatGPT.
\end{minipage}

\bibliographystyle{splncs04}
\bibliography{references}

@misc{UNDissasters,
  title = {{UN report: Dramatic rise in climate disaster over last twenty years}},
  author       = {{United Nations Office for Disaster Risk Reduction}},
  howpublished = {\url{https://www.undrr.org/news/un-report-dramatic-rise-climate-disaster-over-last-twenty-years}},
  note = {Accessed: 2025-11-10}
}

@inproceedings{kim2021graph,
  title={Graph neural network based scene change detection using scene graph embedding with hybrid classification loss},
  author={Kim, Soyeon and Joo, Kyung-no and Youn, Chan-Hyun},
  booktitle={2021 International Conference on Information and Communication Technology Convergence (ICTC)},
  pages={190--195},
  year={2021},
  organization={IEEE}
}

@inproceedings{bandara2022transformer,
  title={A transformer-based siamese network for change detection},
  author={Bandara, Wele Gedara Chaminda and Patel, Vishal M},
  booktitle={IGARSS 2022-2022 IEEE International Geoscience and Remote Sensing Symposium},
  pages={207--210},
  year={2022},
  organization={IEEE}
}

@article{ruuvzivcka2022ravaen,
  title={RaV{\AE}n: unsupervised change detection of extreme events using ML on-board satellites},
  author={R\r{u}{\v{z}}i{\v{c}}ka, V{\'\i}t and Vaughan, Anna and De Martini, Daniele and Fulton, James and Salvatelli, Valentina and Bridges, Chris and Mateo-Garcia, Gonzalo and Zantedeschi, Valentina},
  journal={Scientific reports},
  volume={12},
  number={1},
  pages={16939},
  year={2022},
  publisher={Nature Publishing Group UK London}
}

@article{ding2024adapting,
  title={Adapting segment anything model for change detection in VHR remote sensing images},
  author={Ding, Lei and Zhu, Kun and Peng, Daifeng and Tang, Hao and Yang, Kuiwu and Bruzzone, Lorenzo},
  journal={IEEE Transactions on Geoscience and Remote Sensing},
  volume={62},
  pages={1--11},
  year={2024},
  publisher={IEEE}
}

@article{yadav2024unsupervised,
  title={Unsupervised flood detection on SAR time series using variational autoencoder},
  author={Yadav, Ritu and Nascetti, Andrea and Azizpour, Hossein and Ban, Yifang},
  journal={International Journal of Applied Earth Observation and Geoinformation},
  volume={126},
  pages={103635},
  year={2024},
  publisher={Elsevier}
}

@article{herec2025sttorm,
  title={STTORM-CD: Low-Demand and High-Impact Disaster Monitoring Onboard Satellites Using Change Detection},
  author={Herec, Jon{\'a}{\v{s}} and Sedmidubsk{\`y}, Jan and Pito{\v{n}}{\'a}k, Rado},
  year={2025},
  journal={Research Square preprint:6334392},
}

@article{serief2023deep,
  title={Deep-learning-based system for change detection onboard earth observation small satellites},
  author={Serief, Chahira and Ghelamallah, Youcef and Bentoutou, Youcef},
  journal={IEEE Journal of Selected Topics in Applied Earth Observations and Remote Sensing},
  volume={16},
  pages={8115--8124},
  year={2023},
  publisher={IEEE}
}

@article{zhang2024domain,
  title={Domain knowledge-driven variational recurrent networks for drought monitoring},
  author={Zhang, Mengxue and Fernandez-Torres, Miguel-Angel and Camps-Valls, Gustau},
  journal={Remote Sensing of Environment},
  volume={311},
  pages={114252},
  year={2024},
  publisher={Elsevier}
}

@article{inzerillo2025compress,
  title={Compress-Align-Detect: onboard change detection from unregistered images},
  author={Inzerillo, Gabriele and Valsesia, Diego and Fiengo, Aniello and Magli, Enrico},
  journal={arXiv preprint arXiv:2507.15578},
  year={2025}
}

@inproceedings{xu2025pushing,
  title={Pushing Trade-Off Boundaries: Compact yet Effective Remote Sensing Change Detection},
  author={Xu, Luosheng and Zhang, Dalin and Song, Zhaohui},
  booktitle={Proceedings of the 33rd ACM International Conference on Multimedia},
  pages={641--649},
  year={2025}
}

@article{hafner2025continuous,
  title={Continuous urban change detection from satellite image time series with temporal feature refinement and multi-task integration},
  author={Hafner, Sebastian and Fang, Heng and Azizpour, Hossein and Ban, Yifang},
  journal={IEEE Transactions on Geoscience and Remote Sensing},
  year={2025},
  publisher={IEEE}
}

@article{pang2024exploring,
  title={Exploring model compression limits and laws: A pyramid knowledge distillation framework for satellite-on-orbit object recognition},
  author={Pang, Yanhua and Zhang, Yamin and Wang, Yi and Wei, Xiaofeng and Chen, Bo},
  journal={IEEE Transactions on Geoscience and Remote Sensing},
  volume={62},
  pages={1--13},
  year={2024},
  publisher={IEEE}
}

@article{szwarcman2024prithvi,
  title={Prithvi-eo-2.0: A versatile multi-temporal foundation model for earth observation applications},
  author={Szwarcman, Daniela and Roy, Sujit and Fraccaro, Paolo and G{\'\i}slason, {\TH}orsteinn El{\'\i} and Blumenstiel, Benedikt and Ghosal, Rinki and de Oliveira, Pedro Henrique and Almeida, Joao Lucas de Sousa and Sedona, Rocco and Kang, Yanghui and others},
  journal={arXiv preprint arXiv:2412.02732},
  year={2024}
}

@article{jakubik2025terramind,
  title={Terramind: Large-scale generative multimodality for earth observation},
  author={Jakubik, Johannes and Yang, Felix and Blumenstiel, Benedikt and Scheurer, Erik and Sedona, Rocco and Maurogiovanni, Stefano and Bosmans, Jente and Dionelis, Nikolaos and Marsocci, Valerio and Kopp, Niklas and others},
  journal={arXiv preprint arXiv:2504.11171},
  year={2025}
}

@article{gomes2025terratorch,
  title={TerraTorch: The Geospatial Foundation Models Toolkit},
  author={Gomes, Carlos and Blumenstiel, Benedikt and Almeida, Joao Lucas de Sousa and de Oliveira, Pedro Henrique and Fraccaro, Paolo and Escofet, Francesc Marti and Szwarcman, Daniela and Simumba, Naomi and Kienzler, Romeo and Zadrozny, Bianca},
  journal={arXiv preprint arXiv:2503.20563},
  year={2025}
}

@inproceedings{bastani2023satlaspretrain,
  title={Satlaspretrain: A large-scale dataset for remote sensing image understanding},
  author={Bastani, Favyen and Wolters, Piper and Gupta, Ritwik and Ferdinando, Joe and Kembhavi, Aniruddha},
  booktitle={Proceedings of the IEEE/CVF International Conference on Computer Vision},
  pages={16772--16782},
  year={2023}
}

@article{wang2023ssl4eo,
  title={SSL4EO-S12: A large-scale multimodal, multitemporal dataset for self-supervised learning in Earth observation [Software and Data Sets]},
  author={Wang, Yi and Braham, Nassim Ait Ali and Xiong, Zhitong and Liu, Chenying and Albrecht, Conrad M and Zhu, Xiao Xiang},
  journal={IEEE Geoscience and Remote Sensing Magazine},
  volume={11},
  number={3},
  pages={98--106},
  year={2023},
  publisher={IEEE}
}

@inproceedings{daudt2018fully,
  title={Fully convolutional siamese networks for change detection},
  author={Daudt, Rodrigo Caye and Le Saux, Bertr and Boulch, Alexandre},
  booktitle={2018 25th IEEE international conference on image processing (ICIP)},
  pages={4063--4067},
  year={2018},
  organization={IEEE}
}

@inproceedings{unet,
  title={U-net: Convolutional networks for biomedical image segmentation},
  author={Ronneberger, Olaf and Fischer, Philipp and Brox, Thomas},
  booktitle={International Conference on Medical image computing and computer-assisted intervention},
  pages={234--241},
  year={2015},
  organization={Springer}
}

@article{vaswani2017attention,
  title={Attention is all you need},
  author={Vaswani, Ashish and Shazeer, Noam and Parmar, Niki and Uszkoreit, Jakob and Jones, Llion and Gomez, Aidan N and Kaiser, {\L}ukasz and Polosukhin, Illia},
  journal={Advances in neural information processing systems},
  volume={30},
  year={2017}
}

@inproceedings{kirillov2023segment,
  title={Segment anything},
  author={Kirillov, Alexander and Mintun, Eric and Ravi, Nikhila and Mao, Hanzi and Rolland, Chloe and Gustafson, Laura and Xiao, Tete and Whitehead, Spencer and Berg, Alexander C and Lo, Wan-Yen and others},
  booktitle={Proceedings of the IEEE/CVF international conference on computer vision},
  pages={4015--4026},
  year={2023}
}

@article{zhao2023fast,
  title={Fast segment anything},
  author={Zhao, Xu and Ding, Wenchao and An, Yongqi and Du, Yinglong and Yu, Tao and Li, Min and Tang, Ming and Wang, Jinqiao},
  journal={arXiv preprint arXiv:2306.12156},
  year={2023}
}

@inproceedings{kmeans,
  title={Some methods of classification and analysis of multivariate observations},
  author={McQueen, James B},
  booktitle={Proc. of 5th Berkeley Symposium on Math. Stat. and Prob.},
  pages={281--297},
  year={1967}
}

@article{kingma2013auto,
  title={Auto-encoding variational bayes},
  author={Kingma, Diederik P and Welling, Max},
  journal={arXiv preprint arXiv:1312.6114},
  year={2013}
}

@article{hochreiter1997long,
  title={Long short-term memory},
  author={Hochreiter, Sepp and Schmidhuber, J{\"u}rgen},
  journal={Neural computation},
  volume={9},
  number={8},
  pages={1735--1780},
  year={1997},
  publisher={MIT press}
}

@article{hinton2015distilling,
  title={Distilling the knowledge in a neural network},
  author={Hinton, Geoffrey and Vinyals, Oriol and Dean, Jeff},
  journal={arXiv preprint arXiv:1503.02531},
  year={2015}
}

@inproceedings{sentinel,
  title={Copernicus Sentinel-2 mission: products, algorithms and Cal/Val},
  author={Gascon, Ferran and Cadau, Enrico and Colin, Olivier and Hoersch, Bianca and Isola, Claudia and Fern{\'a}ndez, B L{\'o}pez and Martimort, Philippe},
  booktitle={Earth observing systems XIX},
  volume={9218},
  pages={455--463},
  year={2014},
  organization={SPIE}
}

@article{williams2006landsat,
  title={Landsat},
  author={Williams, Darrel L and Goward, Samuel and Arvidson, Terry},
  journal={Photogrammetric Engineering \& Remote Sensing},
  volume={72},
  number={10},
  pages={1171--1178},
  year={2006},
  publisher={American Society for Photogrammetry and Remote Sensing}
}

@inproceedings{liu2021swin,
  title={Swin transformer: Hierarchical vision transformer using shifted windows},
  author={Liu, Ze and Lin, Yutong and Cao, Yue and Hu, Han and Wei, Yixuan and Zhang, Zheng and Lin, Stephen and Guo, Baining},
  booktitle={Proceedings of the IEEE/CVF international conference on computer vision},
  pages={10012--10022},
  year={2021}
}

@inproceedings{he2016deep,
  title={Deep residual learning for image recognition},
  author={He, Kaiming and Zhang, Xiangyu and Ren, Shaoqing and Sun, Jian},
  booktitle={Proceedings of the IEEE conference on computer vision and pattern recognition},
  pages={770--778},
  year={2016}
}

@article{dosovitskiy2020image,
  title={An image is worth 16x16 words: Transformers for image recognition at scale},
  author={Dosovitskiy, Alexey},
  journal={arXiv preprint arXiv:2010.11929},
  year={2020}
}

@misc{jetson,
  title = {{NVIDIA Jetson Orin: Next-level AI performance for next-gen robotics and edge solutions.}},
  author       = {{NVIDIA}},
  howpublished = {\url{https://www.nvidia.com/en-us/autonomous-machines/embedded-systems/jetson-orin/}},
  note = {Accessed: 2025-11-11}
}

@inproceedings{kirillov2017unified,
  title={A unified architecture for instance and semantic segmentation},
  author={Kirillov, Alexander and He, Kaiming and Girshick, Ross and Doll{\'a}r, Piotr},
  booktitle={Computer Vision and Pattern Recognition Conference},
  year={2017},
  organization={CVPR}
}

@misc{Iakubovskii:2019,
  Author = {Pavel Iakubovskii},
  Title = {Segmentation Models Pytorch},
  Year = {2019},
  Publisher = {GitHub},
  Journal = {GitHub repository},
  Howpublished = {\url{https://github.com/qubvel/segmentation\_models.pytorch}}
}

@inproceedings{yun2019cutmix,
  title={Cutmix: Regularization strategy to train strong classifiers with localizable features},
  author={Yun, Sangdoo and Han, Dongyoon and Oh, Seong Joon and Chun, Sanghyuk and Choe, Junsuk and Yoo, Youngjoon},
  booktitle={Proceedings of the IEEE/CVF international conference on computer vision},
  pages={6023--6032},
  year={2019}
}

@article{loshchilov2017decoupled,
  title={Decoupled weight decay regularization},
  author={Loshchilov, Ilya and Hutter, Frank},
  journal={arXiv preprint arXiv:1711.05101},
  year={2017}
}

@article{loshchilov2016sgdr,
  title={Sgdr: Stochastic gradient descent with warm restarts},
  author={Loshchilov, Ilya and Hutter, Frank},
  journal={arXiv preprint arXiv:1608.03983},
  year={2016}
}

@inproceedings{mao2023cross,
  title={Cross-entropy loss functions: Theoretical analysis and applications},
  author={Mao, Anqi and Mohri, Mehryar and Zhong, Yutao},
  booktitle={International conference on Machine learning},
  pages={23803--23828},
  year={2023},
  organization={pmlr}
}

@inproceedings{sudre2017generalised,
  title={Generalised dice overlap as a deep learning loss function for highly unbalanced segmentations},
  author={Sudre, Carole H and Li, Wenqi and Vercauteren, Tom and Ourselin, Sebastien and Jorge Cardoso, M},
  booktitle={International Workshop on Deep Learning in Medical Image Analysis},
  pages={240--248},
  year={2017},
  organization={Springer}
}

@article{denis2016evolution,
  title={The evolution of Earth Observation satellites in Europe and its impact on the performance of emergency response services},
  author={Denis, Gil and de Boissezon, H{\'e}l{\`e}ne and Hosford, Steven and Pasco, Xavier and Montfort, Bruno and Ranera, Franck},
  journal={Acta Astronautica},
  volume={127},
  pages={619--633},
  year={2016},
  publisher={Elsevier}
}

@article{krizhevsky2012imagenet,
  title={Imagenet classification with deep convolutional neural networks},
  author={Krizhevsky, Alex and Sutskever, Ilya and Hinton, Geoffrey E},
  journal={Advances in neural information processing systems},
  volume={25},
  year={2012}
}

@inproceedings{amit2017disaster,
  title={Disaster detection from aerial imagery with convolutional neural network},
  author={Amit, Siti Nor Khuzaimah Binti and Aoki, Yoshimitsu},
  booktitle={2017 international electronics symposium on knowledge creation and intelligent computing (IES-KCIC)},
  pages={239--245},
  year={2017},
  organization={IEEE}
}

@article{nemni2020fully,
  title={Fully convolutional neural network for rapid flood segmentation in synthetic aperture radar imagery},
  author={Nemni, Edoardo and Bullock, Joseph and Belabbes, Samir and Bromley, Lars},
  journal={Remote Sensing},
  volume={12},
  number={16},
  pages={2532},
  year={2020},
  publisher={MDPI}
}

@inproceedings{ronneberger2015u,
  title={U-net: Convolutional networks for biomedical image segmentation},
  author={Ronneberger, Olaf and Fischer, Philipp and Brox, Thomas},
  booktitle={International Conference on Medical image computing and computer-assisted intervention},
  pages={234--241},
  year={2015},
  organization={Springer}
}

@inproceedings{long2015fully,
  title={Fully convolutional networks for semantic segmentation},
  author={Long, Jonathan and Shelhamer, Evan and Darrell, Trevor},
  booktitle={Proceedings of the IEEE conference on computer vision and pattern recognition},
  pages={3431--3440},
  year={2015}
}

@article{rashkovetsky2021wildfire,
  title={Wildfire detection from multisensor satellite imagery using deep semantic segmentation},
  author={Rashkovetsky, Dmitry and Mauracher, Florian and Langer, Martin and Schmitt, Michael},
  journal={IEEE Journal of Selected Topics in Applied Earth Observations and Remote Sensing},
  volume={14},
  pages={7001--7016},
  year={2021},
  publisher={IEEE}
}

@article{bragagnolo2021convolutional,
  title={Convolutional neural networks applied to semantic segmentation of landslide scars},
  author={Bragagnolo, L and Rezende, LR and Da Silva, RV and Grzybowski, JMV},
  journal={Catena},
  volume={201},
  pages={105189},
  year={2021},
  publisher={Elsevier}
}

@inproceedings{schottl2024real,
  title={Real-time on-orbit fire detection on forest-2},
  author={Sch{\"o}ttl, Fabian and Spichtinger, Andrea and Franquinet, Julian and Langer, Martin},
  booktitle={IGARSS 2024-2024 IEEE International Geoscience and Remote Sensing Symposium},
  pages={2360--2364},
  year={2024},
  organization={IEEE}
}

\newpage
\appendix
\section{Baseline low-quality image identification}\label{app:bad_images}

Table~\ref{tab:time} shows the mean standard deviation of the difference from the best result in the series for bi-temporal baseline. 

\begin{table}[ht]
    \centering
    \caption{Bi-temporal baseline F1 difference to the best result.}
    \label{tab:time}
    
    \begin{tabular}{c|c|c}
\toprule
\multirow{2}{*}{\textbf{Scene}} & \multirow{2}{*}{ \textbf{ Pre-disaster image idx}} & \textbf{F1-score diff  to Maximum } \\
 & & \textbf{mean ± std} \\
\midrule
\multirow{4}{*}{Brazil}  &       1 &  -0.021 ± 0.033                                \\
        &       2 &  -0.007 ± 0.008                                \\
        &       3 &  \textbf{-0.101 ± 0.067} \\
        &       4 &  -0.05 ± 0.057                                 \\
\midrule
\multirow{4}{*}{Germany} &       1 &  -0.077 ± 0.048                                \\
        &       2 &  -0.005 ± 0.008                                \\
        &       3 &  \textbf{-0.118 ± 0.065} \\
        &       4 &  -0.071 ± 0.054                                \\
\midrule
\multirow{4}{*}{Laos}    &       1 &  -0.01 ± 0.012                                 \\
        &       2 &  -0.029 ± 0.01                                 \\
        &       3 &  -0.008 ± 0.012                                \\
        &       4 &  -0.018 ± 0.015                                \\
\midrule
\multirow{4}{*}{Niger}   &       1 &  -0.0 ± 0.001                                  \\
        &       2 &  -0.0 ± 0.0                                    \\
        &       3 &  -0.0 ± 0.001                                  \\
        &       4 &  -0.001 ± 0.002                                \\
\bottomrule
\end{tabular}
\end{table}

\end{document}